\documentclass[letterpaper, 10 pt, conference]{ieeeconf} 
\IEEEoverridecommandlockouts                              
\usepackage{graphicx}      
\usepackage{amsmath}       
\usepackage{amssymb}       
\usepackage{booktabs}
\usepackage{cite}
\usepackage{algorithm}
\usepackage{algorithmic}

\title{\LARGE \bf
MedVLA: A Hierarchical Vision-Language-Action Framework for Closed-Loop Precision Medical Robot Manipulation
}

\author{Junjie Xie$^{\dagger}$, Angen Ye$^{\dagger}$, Chuxuan He, Yujia Song, and Dapeng Zhang$^{*}$
\thanks{This work is supported by the Beijing Natural Science Foundation Haidian Original Innovation Joint Fund Project (L232035, L222154).}%
\thanks{$^{\dagger}$Junjie Xie and Angen Ye contributed equally to this work.}%
\thanks{Junjie Xie, Angen Ye, Yujia Song, and Dapeng Zhang are with the Institute of Automation, Chinese Academy of Sciences, Beijing 100190, China. Chuxuan He is with Zhejiang Gongshang University, Hangzhou 310018, China. $^{*}$Corresponding author: Dapeng Zhang (e-mail: {\tt\small dapeng.zhang@ia.ac.cn}).}%
}

\begin{document}

\maketitle
\thispagestyle{empty}
\pagestyle{empty}

\begin{abstract}
\textbf{Precision medical robotics demands adaptive decision-making under strict safety, interpretability, and execution constraints. Although recent Vision-Language-Action (VLA) models show strong multimodal reasoning ability, their continuous action generation paradigm is not well suited for precision medical tasks, where reliable closed-loop operation may also depend on non-action system function calls. To address this gap, we propose MedVLA, a hierarchical framework that couples high-level multimodal reasoning with low-level function-constrained execution. We further introduce a scalable multi-agent pipeline to generate skill-oriented chain-of-thought(CoT) data for structured training. Built on different multimodal large-model backbones, MedVLA consistently improves performance after fine-tuning, demonstrating the effectiveness of the proposed framework across model variants. Under identical initial conditions, we perform 100 closed-loop flexible electrode implantation trials. The results show that MedVLA achieves a 95.0\% task success rate, substantially outperforming representative VLA baselines, including OpenVLA (8\%) and $\pi_0$ (15\%), in accuracy, stability, and safety. These results indicate that structured reasoning with constrained function-level execution is a practical route toward deployable precision medical robotics.}
\end{abstract}




\section{INTRODUCTION}

Precision medical robotics has become an important research direction at the intersection of medical engineering, robotics, and intelligent systems \cite{iakovidis2025medical}. Its goal is not merely to complete predefined operations, but to maintain stable control of instruments and tissues in medical environments characterized by high risk, strong constraints, and extremely low fault tolerance. Unlike general manipulation in open environments, precision medical tasks are typically performed in narrow surgical fields and require continuous perception, real-time judgment, and dynamic correction within millimeter-scale or smaller workspaces \cite{wu2024review, liu2024evolution, dagnino2024robot, huang2023mri}. In scenarios such as minimally invasive intervention, neural modulation, and electrode implantation, the robot must not only localize key targets from visual observations, but also continuously update subsequent actions by jointly considering instrument state, task phase, and operation goals, thereby enabling safe and accurate closed-loop execution under continuous feedback \cite{dagnino2024robot, huang2023mri, maier2017surgical}. Such settings therefore require integrated high-precision decision-making, constrained execution, and closed-loop adjustment under complex operational constraints \cite{huang2023mri, maier2017surgical}.

Medical robotics has made sustained progress in mechanism design, visual navigation, motion control, and human--robot collaboration \cite{liu2024evolution, dagnino2024robot, moustris2011evolution, rivas2021review}. However, most existing systems still remain within the ``human-led, robot-assisted'' paradigm, where intelligence is reflected in local functional enhancement rather than autonomous decision-making throughout the task process \cite{lee2024levels, varghese2024artificial, vasey2023intraoperative, kenig2024artificial}. Meanwhile, Vision-Language-Action (VLA) models such as OpenVLA and $\pi_0$ have unified visual perception, language understanding, and action generation within a single framework, showing strong cross-task generalization in open-world manipulation \cite{kim2024openvla, black2024pi_0}. Yet a substantial gap remains between such general-purpose VLA paradigms and precision medical requirements. Schmidgall \textit{et al.} pointed out that robot-assisted surgery faces distinctive challenges including limited open surgical data, soft-tissue modeling difficulty, and stricter safety validation needs \cite{schmidgall2024general}. Lee \textit{et al.} confirmed through a review of FDA-cleared surgical robots that most deployed systems operate at low autonomy levels \cite{lee2024levels}. Zargarzadeh \textit{et al.} further showed that in surgical settings, higher-level reasoning must be explicitly coordinated with lower-level execution rather than left to unconstrained end-to-end action generation \cite{zargarzadeh2025decision}. These findings indicate that directly applying general-purpose VLA models to precision medical tasks remains insufficient in execution accuracy, safety constraints, and behavioral determinism.
\begin{figure*}[t]
  \centering
  \includegraphics[width=\textwidth]{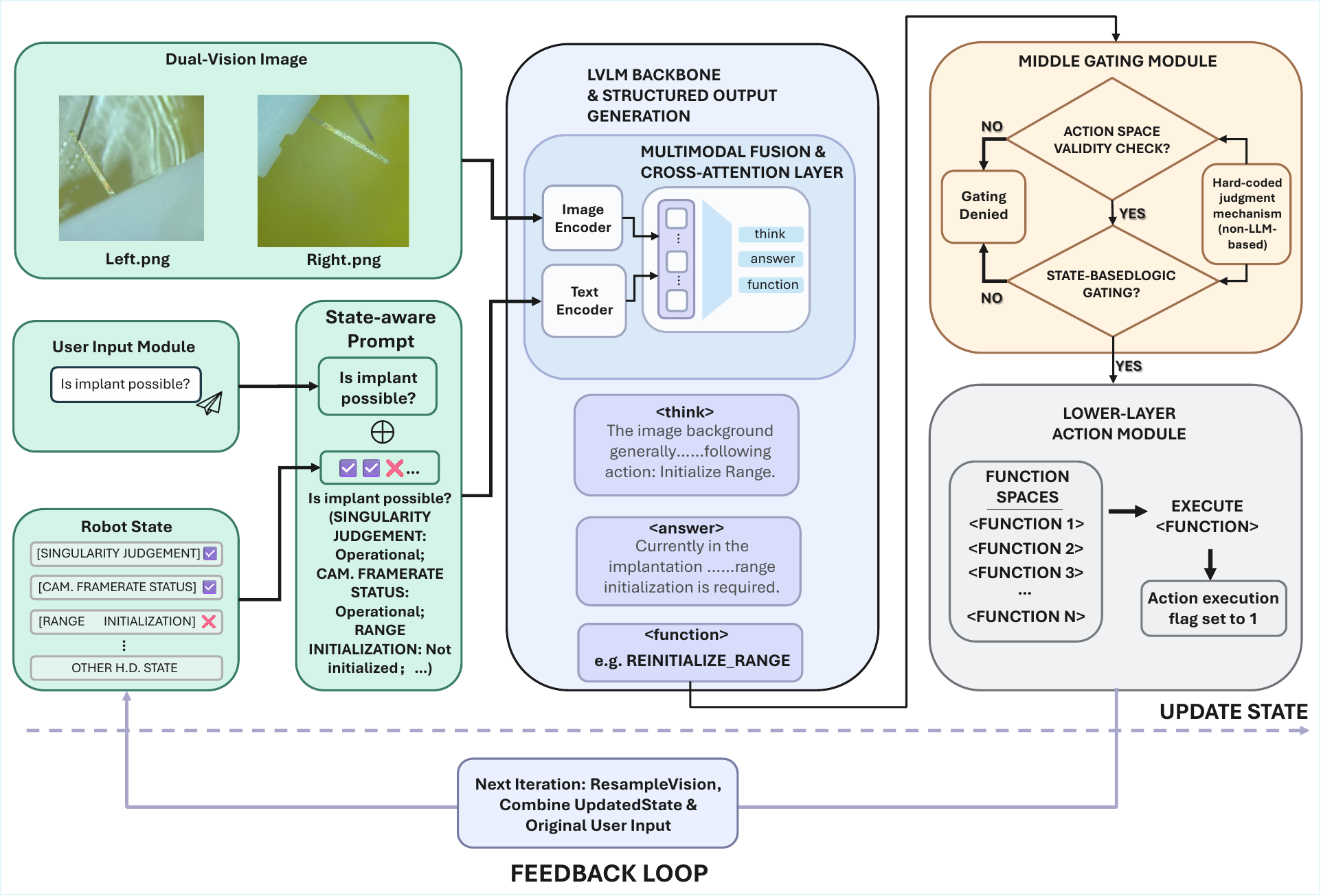} 
  \caption{Overview of the proposed hierarchical closed-loop framework for precision medical robot operation. Dual-view images, user instruction, and robot state are combined into a state-aware prompt and processed by the LVLM to generate structured outputs (\texttt{think}, \texttt{answer}, and \texttt{function}). The predicted function is then verified by the middle gating module through action-space validity checking and state-based logic gating before being executed by the lower-layer action module. Updated visual observations and robot states are fed back into the next iteration to form a closed-loop process.}
  \label{fig1}
\end{figure*}
Beyond the execution-level mismatch, data acquisition has become another major bottleneck. Previous studies have relied on small-scale experimental data or task-specific samples collected on dedicated platforms, which remain insufficient for systems that must learn high-level reasoning and closed-loop feedback mechanisms \cite{maier2017surgical, schmidgall2024general, li2024deep, maier2022surgical}. Data collection for precision medical tasks is inherently costly due to specialized platforms, dense annotation, and expert participation \cite{rivas2021review, kenig2024artificial, sone2023evolution}. Moreover, recent work shows that raw visual input with direct supervision cannot capture the full decision chain of closed-loop medical operation; richer representations such as hierarchical phase--task--action annotations \cite{bilal2025temset}, explicit phase recognition \cite{kolbinger2024artificial}, and structured surgical workflow modeling \cite{li2024deep, wagner2023comparative, li2025surgical} are necessary. What is therefore needed is a scalable data construction mechanism that can generate structured task data efficiently while remaining extensible to new precision medical scenarios \cite{maier2017surgical, maier2022surgical}.

To address these issues, this paper proposes \textbf{MedVLA}, a hierarchical vision-language-action framework for closed-loop precision medical robot manipulation. While hierarchical control and skill-based planning have been explored in general robotics \cite{kim2024openvla, black2024pi_0}, MedVLA differs from prior work in three key respects. First, unlike high-level planner / low-level controller frameworks that decompose tasks into language-level subgoals and delegate execution to learned continuous policies, MedVLA outputs structured function calls verified by an explicit state-based gating module before reaching the executor, making each decision auditable and constrained rather than open-ended. Second, the gating mechanism enforces domain-specific preconditions tied to the physical task state---not merely action-space membership---so that unsafe or premature operations are blocked before execution. Third, the framework integrates chain-of-thought reasoning into the decision triplet, coupling interpretable reasoning traces with function prediction to support both post-hoc analysis and improved intermediate judgment accuracy. Together, these design choices address the safety, determinism, and verifiability requirements that distinguish precision medical manipulation from open-world robotic tasks.

Specifically, the upper module is centered on a fine-tuned large model that integrates visual observations, system states, and task goals to produce structured decision outputs. The lower module invokes constrained skill interfaces according to these decisions, executes them in a single-step manner, updates the environment state after each operation, and feeds the updated state back to the upper module, thereby forming a progressively advancing closed-loop process. Through this design, MedVLA preserves the high-level semantic understanding and task reasoning ability of large models \cite{kim2024openvla, black2024pi_0} while strictly restricting actual execution to a constrained and verifiable skill space, making it better suited to the safety, determinism, and execution-precision requirements of precision medical tasks \cite{iakovidis2025medical, dagnino2024robot}. In addition, to address the high cost and limited scalability of real-world medical data collection, we design a skill-oriented multi-agent data generation mechanism for MedVLA. Multiple agents collaboratively construct instruction, state, decision, and skill-invocation data throughout the task process, while a cosine-similarity-based deduplication strategy removes highly repetitive samples. This process yields a structured dataset with lower redundancy, stronger generalization ability, and clearer organization for MedVLA training and deployment.

We conduct systematic validation of the proposed method on a real implantable electrode operation platform. The primary contributions of this work are summarized as follows:

\textbf{1)} We propose a hierarchical architecture tailored for precision medical tasks, which achieves safe closed-loop control over a constrained skill space by decoupling high-level judgment from low-level execution.

\textbf{2)} We design a skill-oriented multi-agent data generation mechanism and construct a structured dataset of approximately 30,000 samples with low redundancy and strong generalization, providing an effective data foundation for closed-loop decision learning.

\textbf{3)} We perform system-level validation on a real platform and show that the proposed method achieves a 95.0\% success rate in complete closed-loop tasks, outperforming OpenVLA and $\pi_0$ by 87.0 and 80.0 percentage points, respectively, thereby demonstrating its effectiveness in balancing precision execution and intelligent decision-making.


\section{METHODOLOGY}

\subsection{Task Formulation and Problem Modeling}

Precision medical robotic tasks are structured decision-making problems under multimodal and state constraints, rather than direct regressions from visual observations to continuous actions. Compared with general manipulation, they involve stringent precision requirements and explicit execution stages. At each step, the system decision depends on visual observations, execution progress, completed actions, and task objectives. We therefore formulate the task as an iterative process centered on state-constrained function calls.

At time $t$, the system receives binocular image observations $I_t=\{I_t^{(1)}, I_t^{(2)}\}$, user task input $u_t$, and internal robot state $s_t$. Here, $I_t$ captures the surgical scene and local geometric relations, $u_t$ specifies the task intent, and $s_t$ records the execution stage and internal decision variables. Based on these inputs, the system predicts a structured decision triplet
\begin{equation}
y_t=\{\langle\text{think}\rangle_t, \langle\text{answer}\rangle_t, \langle\text{function}\rangle_t\},
\end{equation}
where $\langle\text{think}\rangle_t$ denotes the reasoning process, $\langle\text{answer}\rangle_t$ the intermediate judgment, and $\langle\text{function}\rangle_t$ the next function call. The core mapping learned in this work is
\begin{equation}
(I_t, u_t, s_t) \to y_t,
\end{equation}
rather than the conventional mapping $(I_t, u_t) \to a_t$.

Let $\mathcal{F}$ denote the discrete executable function space, and let $\mathcal{F}(s_t)\subseteq\mathcal{F}$ denote the subset allowed under the current state. The predicted function must satisfy
\begin{equation}
\langle\text{function}\rangle_t \in \mathcal{F}(s_t).
\end{equation}
The overall task is thus formulated as the closed-loop process
\begin{equation}
(I_t, u_t, s_t) \to y_t \to \langle\text{function}\rangle_t \to (s_{t+1}, I_{t+1}),
\end{equation}
which repeats until termination.

\begin{figure*}[t]
  \centering
  \includegraphics[width=0.95\textwidth]{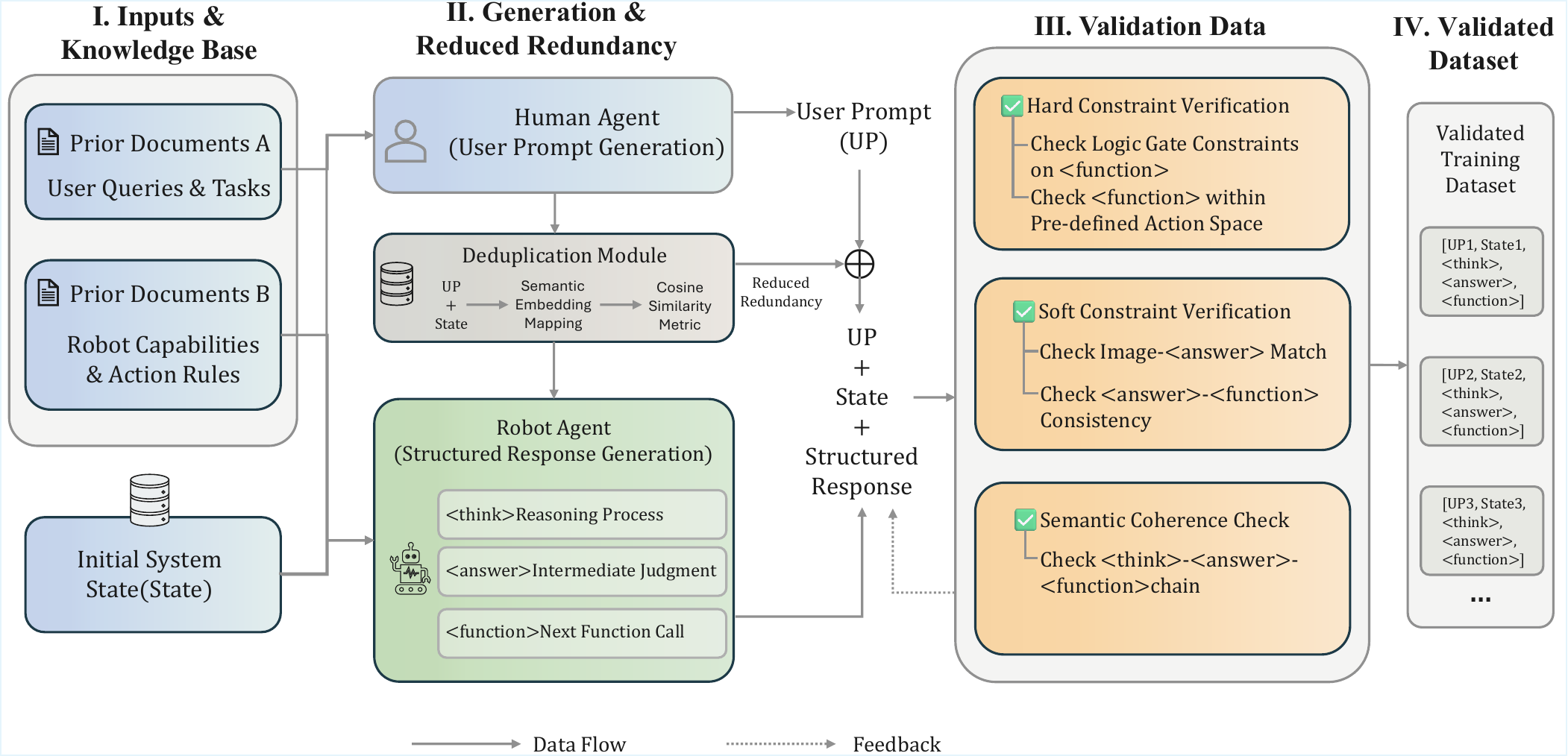}
  \caption{Structured data generation pipeline based on multi-agent collaboration. Prior documents and initial system states are used to generate user prompts. The combined prompt-state inputs are embedded using the hidden representations of the LVLM backbone, and redundant samples are removed by cosine similarity. The retained samples are then used to generate structured responses (\texttt{think}, \texttt{answer}, and \texttt{function}) and further filtered by hard constraints, soft constraints, and semantic coherence checks to form the validated training dataset.}
  \label{fig:data_gen}
\end{figure*}

\subsection{Hierarchical Closed-Loop Control Framework}

To ensure stable execution under stringent constraints, we design a hierarchical closed-loop framework that organizes perception, reasoning, verification, and execution into a unified workflow. As shown in Fig.~1, the framework consists of multimodal perception and state inputs, a high-level decision module, a middle gating module, a low-level action module, and a feedback update loop.

At each step, the robot state is explicitly combined with the user instruction to form a state-aware prompt. This allows the model to reason not only about the visible scene, but also about what is currently executable under the system state. The high-level decision module uses a fine-tuned Large Vision-Language Model (LVLM) to map the image observations and state-aware prompt to the structured triplet $y_t$ defined in Eq.~(1), making the decision process explicit and interpretable.

To prevent invalid or unsafe execution, the predicted $\langle\text{function}\rangle_t$ is checked by the middle gating module, which performs two sequential checks: (i)~\textit{action-space validity}, ensuring that the predicted function belongs to the predefined executable set~$\mathcal{F}$; and (ii)~\textit{state-based logic gating}, verifying that the function is consistent with the current robot state~$s_t$ and satisfies domain-specific preconditions. In the implantation task, these preconditions include geometric alignment status between the needle and electrode hole, vision-system readiness (exposure and focus within operational bounds), and workspace initialization state. Only functions that pass both checks are forwarded to the lower layer for execution.

When the gate rejects a predicted function, the system immediately triggers \texttt{E\_STOP}, initiates a hardware self-check routine, and activates an audible alarm. Given the safety-critical nature of precision surgical tasks, the system then transfers control to the human operator rather than attempting autonomous recovery to an uncertain state. This conservative design reflects the principle that, in high-stakes medical settings, a safe halt with human handover is preferable to autonomous re-planning whose safety cannot be fully guaranteed.

After execution, the system updates the robot state and visual observations, yielding $(s_{t+1}, I_{t+1})$ for the next iteration. This forms an iterative ``Perceive--Reason--Gate--Execute--Update'' loop, summarized in Algorithm~\ref{alg:loop}.

\begin{algorithm}[t]
\caption{Hierarchical Closed-Loop Control}
\label{alg:loop}
\small
\begin{algorithmic}[1]
\REQUIRE Images $I_0$, user input $u$, initial state $s_0$, LVLM $\mathcal{M}$, function space $\mathcal{F}$
\STATE $t \leftarrow 0$
\WHILE{task not terminated}
  \STATE $y_t \leftarrow \mathcal{M}(I_t, u, s_t)$ \COMMENT{Reason: predict triplet}
  \STATE $f_t \leftarrow \langle\text{function}\rangle_t$
  \IF{$f_t \notin \mathcal{F}$ \textbf{or} $f_t \notin \mathcal{F}(s_t)$}
    \STATE Execute \texttt{E\_STOP}; self-check; alarm; hand over \COMMENT{Gate: reject}
  \ENDIF
  \STATE Execute $f_t$; update $(s_{t+1}, I_{t+1})$ \COMMENT{Execute \& Update}
  \STATE $t \leftarrow t + 1$
\ENDWHILE
\end{algorithmic}
\end{algorithm}

\subsection{Structured Data Generation via Multi-Agent Collaboration}

To align the training data with the proposed framework, we construct a structured data generation pipeline based on multi-agent collaboration, redundancy reduction, and constraint verification, as illustrated in Fig.~2.

We use prior task documents, robot capability documents, action rules, and initial system states as knowledge sources for data generation. In this pipeline, we define two LLM-based roles: the \textit{Human Agent}, which simulates a user by generating diverse task prompts from the source materials, and the \textit{Robot Agent}, which produces structured responses given the generated prompts and system states. The term \textit{skill-oriented} refers to the fact that each training sample is organized around a single executable function call (i.e., a predefined robot skill) rather than free-form text or raw action trajectories. The Human Agent first generates user prompts to simulate realistic task requests. To reduce redundancy, each prompt-state pair is embedded using the internal hidden representations of Qwen2.5-VL-3B-Instruct, and highly similar samples are removed by cosine similarity thresholding.

The retained prompt-state pairs are then fed to the Robot Agent, which generates a structured response following the same triplet format as Eq.~(1). The generated response is paired with the corresponding prompt and state for verification. The Verification Agent then performs three levels of checking: hard constraint verification, which tests whether the predicted function satisfies logic-gate constraints and remains within the predefined action space; soft constraint verification, which checks consistency between image observations and $\langle\text{answer}\rangle$, as well as alignment between $\langle\text{answer}\rangle$ and $\langle\text{function}\rangle$; and semantic coherence checking over the full $\langle\text{think}\rangle$--$\langle\text{answer}\rangle$--$\langle\text{function}\rangle$ chain. Only samples that pass all checks are retained to form the final validated training dataset.

This pipeline improves data quality by jointly reducing redundancy and enforcing constraint consistency, producing training data better aligned with the state-aware, function-constrained, and closed-loop characteristics of precision medical robotic tasks.

\section{EXPERIMENTS}

\subsection{Experimental Platform, Setup, and Evaluation Protocol}

\begin{figure}[t]
    \centering
    \includegraphics[width=\columnwidth]{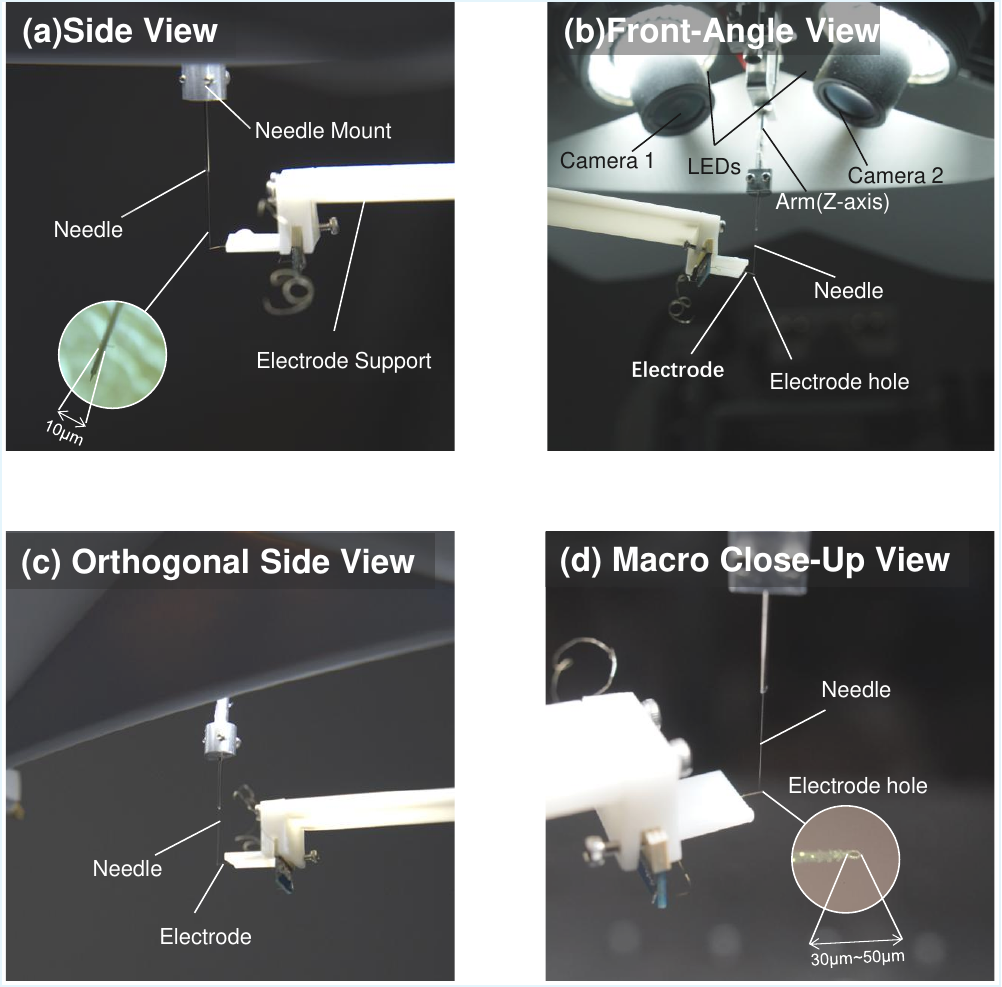}
    \caption{Experimental platform for flexible electrode implantation. (a) Side view of the needle mount, insertion needle, and electrode support, with a zoom-in of the $\sim$10~$\mu$m needle tip. (b) Front-angle view of the overall setup, including the binocular cameras, LEDs, robotic arm, and flexible electrode. (c) Orthogonal side view of the needle--electrode configuration. (d) Macro close-up of the insertion region, showing alignment between the needle and the approximately $30 \times 50$~$\mu$m electrode hole.}
    \label{fig:exp_setup}
\end{figure}

We evaluate the proposed framework on a real-world flexible electrode implantation platform for precision neurosurgical manipulation, where a tungsten insertion needle (${\sim}$10~$\mu$m tip) must pass through a $30 \times 50$~$\mu$m electrode hole (Fig.~\ref{fig:exp_setup}). The micrometer-scale geometric constraints make this task representative of precision surgical robotic systems. We conduct both offline function prediction and real-robot closed-loop evaluation.

The decision space consists of seven discrete functions, grouped into image regulation, state management, safety control, and action execution, as listed in Table~\ref{tab:functions}. Among them, \texttt{ALIGN\_NEEDLE} and \texttt{INSERT} are the two execution-critical functions directly associated with implantation, while the others provide visual adjustment, system recovery, and safe interruption during closed-loop operation.

\begin{table}[t]
\caption{Discrete function space used in the proposed framework. Symbols indicate function categories: $\blacktriangle$ image regulation, $\star$ state management, $\blacksquare$ safety control, and $\bullet$ action execution.}
\label{tab:functions}
\centering
\small
\setlength{\tabcolsep}{3pt}
\begin{tabular}{p{4cm}p{4.2cm}}
\toprule
\textbf{Function} & \textbf{Description} \\
\midrule
$\blacktriangle$ \texttt{ADJUST\_EXPOSURE} & Adjust image exposure to improve brightness and visibility. \\
$\blacktriangle$ \texttt{ADJUST\_FOCUS} & Adjust focus to improve scene sharpness. \\
$\star$ \texttt{REINITIALIZE\_RANGE} & Reset the robot working range when the current state is unsuitable. \\
$\star$ \texttt{RESTART\_CAMERA} & Restart the vision system after camera failure or abnormal observation. \\
$\blacksquare$ \texttt{E\_STOP} & Immediately stop execution for safety protection. \\
$\bullet$ \texttt{ALIGN\_NEEDLE} & Refine the alignment between the insertion needle and the target. \\
$\bullet$ \texttt{INSERT} & Execute the final insertion toward the target position. \\
\bottomrule
\end{tabular}
\end{table}

We organize the evaluation into offline function prediction and real-robot closed-loop execution. For dataset construction, we collect approximately 3,000 real-world samples (synchronized binocular images, function labels, and trajectories at 5~Hz) and expand to ${\sim}$30,000 samples through controlled image perturbations. Combined with the data generation pipeline in Section~II, this provides a unified training foundation for all methods. To verify backbone generality, we fine-tune InternVL2-2B, Qwen2.5-VL-3B, and Llama-3.2-11B-Vision-Instruct. For comparison, we evaluate OpenVLA-7B and $\pi_0$-3.3B in real-robot closed-loop execution. Qwen2.5-VL-3B is selected for deployment due to its favorable trade-off between accuracy and efficiency. We report \textbf{Exact Match (EM)} and \textbf{Macro-F1} for offline evaluation, and closed-loop insertion success rate for real-robot evaluation.

\begin{table*}[t]
\caption{Offline Function Prediction Performance of Different Backbones and Ablation Variants}
\label{table_offline}
\centering
\resizebox{\textwidth}{!}{
\begin{tabular}{lccccccc}
\toprule
\multicolumn{8}{c}{\textbf{Overall Performance on the Action-Balanced Test Set}} \\
\midrule
\textbf{Type} & \textbf{Model} & \textbf{Scale} & \textbf{FT} & \textbf{CoT} & \textbf{EM (\%)} & \textbf{Macro-F1} \\
\midrule
Zero-shot & Qwen2.5-VL & 3B & No & No & 29.60 & 0.1594 \\
Zero-shot & Qwen3-VL & 30B-A3B & No & No & 65.40 & 0.7474 \\
Cross-backbone & InternVL2 & 2B & Yes & Yes & 94.00 & 0.9400 \\
Cross-backbone & Llama-3.2-11B & 11B & Yes & Yes & 95.70 & 0.9570 \\
Ours & Qwen2.5-VL & 3B & Yes & No & 85.30 & 0.8693 \\
\textbf{Ours} & \textbf{Qwen2.5-VL} & \textbf{3B} & \textbf{Yes} & \textbf{Yes} & \textbf{95.00} & \textbf{0.9604} \\
\midrule
\midrule
\multicolumn{8}{c}{\textbf{Per-Function Accuracy (\%)}} \\
\midrule
\textbf{Function (GT)} & \textbf{Support} & \textbf{Qwen2.5-3B} & \textbf{Qwen3-30B} & \textbf{InternVL2-2B} & \textbf{Llama-3.2-11B} & \textbf{Ours w/o CoT} & \textbf{Ours Full} \\
\midrule
ADJUST\_EXPOSURE    & 243 & 100.00 & 84.36 & 99.18 & 100.00 & 99.18 & \textbf{100.00} \\
ADJUST\_FOCUS       & 243 & 0.00   & 20.99 & 95.88 & \textbf{97.12} & 65.84 & 96.71 \\
REINITIALIZE\_RANGE & 141 & 0.00   & 100.00 & 100.00 & 100.00 & 100.00 & 100.00 \\
ALIGN\_NEEDLE       & 113 & 0.00   & 39.82 & 70.80 & \textbf{76.11} & 48.67 & 72.57 \\
RESTART\_CAMERA     & 100 & 1.00   & 96.00 & 100.00 & 100.00 & 100.00 & 100.00 \\
E\_STOP             & 100 & 52.00  & 100.00 & 100.00 & 100.00 & 100.00 & 100.00 \\
INSERT              & 60  & 0.00   & 26.67 & 75.00 & 85.00 & \textbf{93.33} & 81.67 \\
\midrule
\textbf{OVERALL}    & \textbf{1000} & \textbf{29.60} & \textbf{65.40} & \textbf{94.00} & \textbf{95.70} & \textbf{85.30} & \textbf{95.00} \\
\bottomrule
\end{tabular}}
\end{table*}

\subsection{Offline Function Prediction Results}

Table~\ref{table_offline} reports offline function prediction results under zero-shot evaluation, cross-backbone fine-tuning (i.e., applying the same structured training procedure to LVLM backbones other than the primary Qwen2.5-VL), and CoT ablation. Overall, the full Qwen2.5-VL-3B variant reaches 95.00\% accuracy and 0.9604 Macro-F1, substantially outperforming the zero-shot baselines, which supports the effectiveness of structured function-level supervision for precision surgical decision making.

Comparing \textit{Ours w/o CoT} and \textit{Ours Full} shows that the effect of CoT supervision is class-dependent. The largest gains appear on \texttt{ADJUST\_FOCUS} (65.84\% $\rightarrow$ 96.71\%) and \texttt{ALIGN\_NEEDLE} (48.67\% $\rightarrow$ 72.57\%), indicating that CoT is particularly helpful for functions requiring finer judgment of visual quality and intermediate surgical states. By contrast, \texttt{INSERT} decreases from 93.33\% to 81.67\%. Error analysis on the 60 \texttt{INSERT} test samples reveals that the additional false negatives introduced by CoT are predominantly misclassified as \texttt{ALIGN\_NEEDLE}, meaning the model favors further alignment refinement rather than committing to insertion when the reasoning chain expresses residual uncertainty about geometric readiness. This conservative shift is not necessarily harmful: as shown in Section~III-C, permitting additional alignment rounds in closed-loop execution leads to a higher overall insertion success rate (95\% vs.\ 73\% at 3 rounds), because the real-world cost of a redundant \texttt{ALIGN\_NEEDLE} call is far lower than the cost of a premature \texttt{INSERT}. Despite this per-class reduction, CoT improves reasoning over ambiguous intermediate states and raises overall accuracy from 85.30\% to 95.00\%.

Cross-backbone results further show that the framework generalizes well across different LVLM architectures. Under the same structured function-calling setting, both InternVL2-2B and Llama-3.2-11B achieve competitive results, indicating that the framework does not depend on a specific backbone design. The Llama-based variant attains the best overall accuracy, likely due to its larger parameter scale. However, Qwen2.5-VL-3B remains highly competitive with a much smaller model size and offers a better practical trade-off for deployment, and is therefore selected for real-robot experiments.

As shown in Table~\ref{table_offline}, the remaining errors of the full model are mainly concentrated in three functions: \texttt{ADJUST\_FOCUS}, \texttt{ALIGN\_NEEDLE}, and \texttt{INSERT}. By contrast, \texttt{ADJUST\_EXPOSURE}, \texttt{REINITIALIZE\_RANGE}, \texttt{RESTART\_CAMERA}, and \texttt{E\_STOP} are predicted with near-perfect or perfect accuracy, indicating that the major difficulty lies in visually ambiguous intermediate judgments and precision-critical execution decisions rather than recovery or safety-related functions.

\subsection{Closed-Loop Comparison in Real Execution}

We evaluate all methods under a unified 100-trial closed-loop protocol. In each trial, the insertion needle is randomly initialized, and successful insertion is the final criterion. For each initialization, the VLA baseline runs first with all motor trajectories recorded; the robot is then restored to the same start state for our method, ensuring a consistent comparison.
\begin{figure*}[t]
  \centering
  \includegraphics[width=\textwidth]{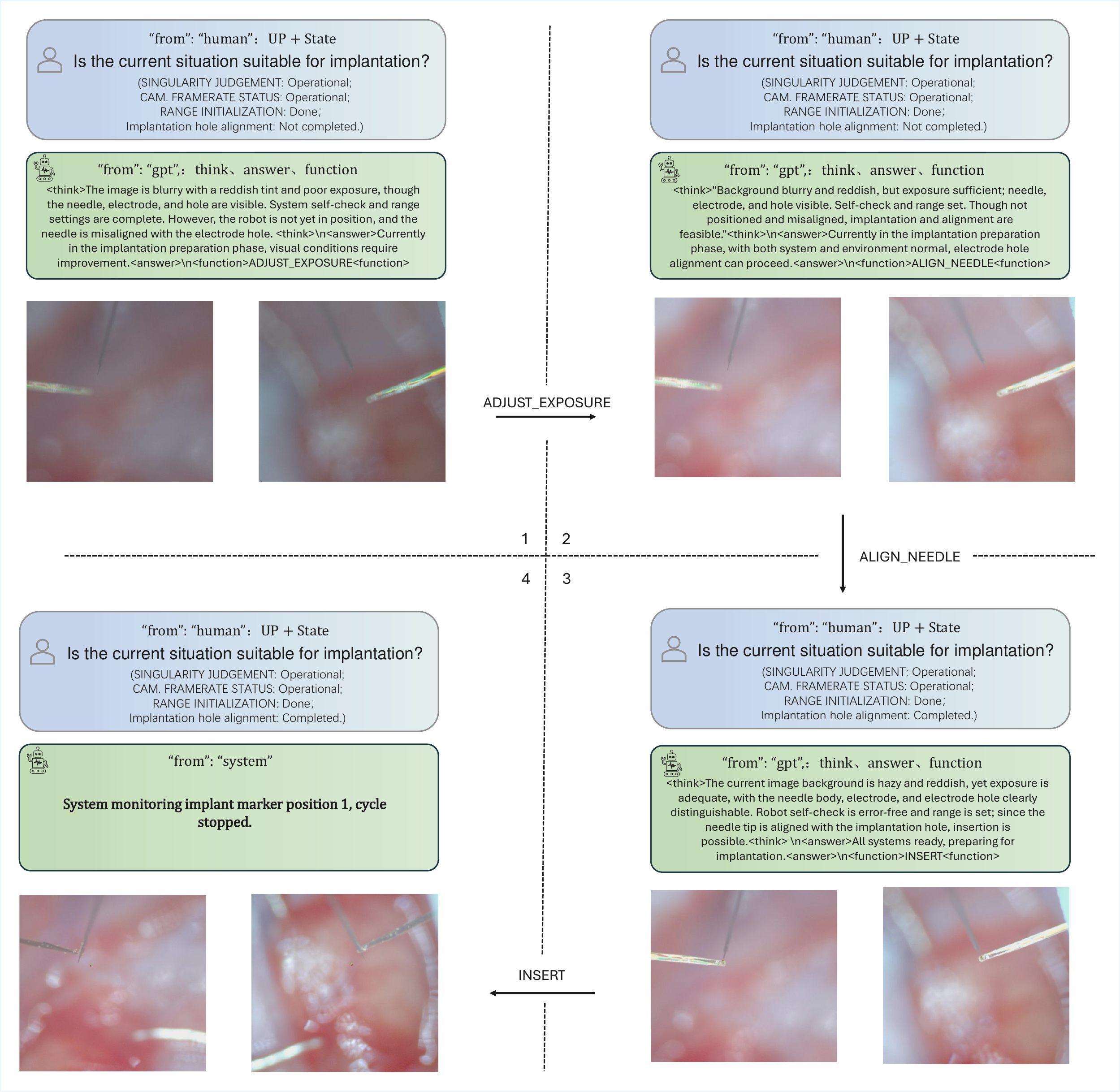}
  \caption{Real-world closed-loop deployment example of the proposed hierarchical framework using the fine-tuned Qwen2.5-VL-3B-Instruct model. Starting from binocular microscopic observations and task prompts, the system sequentially performs \texttt{ADJUST\_EXPOSURE}, \texttt{ALIGN\_NEEDLE}, and \texttt{INSERT}, ultimately completing electrode-hole insertion through structured function-level decision making.}
  \label{fig:trajectories}
\end{figure*}
Under this protocol, OpenVLA achieves \textbf{8} successful insertions out of 100 trials, corresponding to \textbf{8.00\%}, while $\pi_0$ achieves \textbf{15} successful insertions, corresponding to \textbf{15.00\%}. In contrast, the proposed hierarchical framework performs substantially better. With the maximum number of decision rounds limited to \textbf{3}, our method achieves \textbf{73.00\%}; increasing the limit to \textbf{6} further improves the success rate to \textbf{95.00\%}. This result is consistent with the offline observation that the model is more conservative before committing to the final \texttt{INSERT} action. Although such conservativeness may slightly reduce offline \texttt{INSERT} accuracy, it appears to avoid premature or risky insertion attempts in real execution, thereby improving overall success in high-precision electrode implantation.

We chose OpenVLA and $\pi_0$ as baselines because they represent the two dominant VLA paradigms---autoregressive action tokenization and flow-matching continuous control---and comparing against them isolates the benefit of hierarchical function-level execution over end-to-end continuous action generation. Conventional visual-servoing or PID/MPC controllers could serve as domain-specific baselines but are limited to low-level subtasks and lack the high-level multimodal reasoning that the proposed framework targets; a fair comparison would require designing equivalent task-level decision logic around them, which we leave for future work.

\textbf{Inference latency.} In our deployment configuration (Qwen2.5-VL-3B on a single GPU), the average end-to-end latency per decision cycle is approximately 10~seconds, comprising ${\sim}$8~s for LVLM inference (including image encoding and structured output generation), ${\sim}$2~s for state update and gating verification, and negligible time for low-level motor command execution. We note that the 5~Hz in Section~III-A is the offline trajectory-sampling rate used to build dense training data, and is distinct from the deployment decision rate: at runtime the system issues one structured decision per ${\sim}$10~s cycle (${\sim}$0.1~Hz), and the executed function moves the needle by a single constrained step before the next observation is taken. Since the implantation task operates quasi-statically with no moving tissue, this decision rate is acceptable for the current application. For time-critical scenarios, latency could be reduced via distillation, quantization, or speculative decoding.

\textbf{Gating module analysis.} To assess the role of the middle gating module, we conducted over 500 inference trials on the test set under varied prompt formulations. Across all trials, the gate rejected the predicted function only twice, both due to the LVLM generating outputs outside the predefined seven-function action space. In each case, the system immediately triggered \texttt{E\_STOP}, executed a hardware self-check, and activated an audible alarm before handing control to the human operator. The low rejection rate indicates that the fine-tuned LVLM has learned to produce well-formed function calls with high reliability. Nevertheless, the gating module remains essential as a hard safety boundary: because removing it would allow arbitrary model outputs to be directly executed on micrometer-scale surgical hardware, a full ablation (i.e., disabling the gate during physical execution) was not conducted to avoid potential equipment damage. The gating module thus serves as a necessary safeguard whose value lies not in frequent activation but in guaranteeing safe termination under rare but critical failure modes.

\subsection{Real-World Closed-Loop Case Study}

Fig.~\ref{fig:trajectories} shows a representative deployment case using the fine-tuned Qwen2.5-VL-3B-Instruct model. Starting from binocular microscopic observations, the system sequentially issues \texttt{ADJUST\_EXPOSURE}, \texttt{ALIGN\_NEEDLE}, and \texttt{INSERT}, completing electrode-hole insertion through structured closed-loop decision making.
\section{Conclusion}

This paper presents a hierarchical vision-language-action framework for precision medical robotics and validates it on flexible electrode implantation under micrometer-level constraints. By decoupling high-level multimodal decision-making from constrained low-level execution, the proposed method achieves a 95\% closed-loop success rate, addressing a key limitation of continuous-action VLA paradigms in safety-critical settings. The multi-agent data pipeline further provides scalability: new task actions can be supported by extending the pipeline and fine-tuning.

\textbf{Limitations.} The current evaluation is conducted on a single precision medical task with training data generated by the proposed pipeline, which may favor the framework by construction. Validation on independent datasets, alternative surgical tasks, and data collected by external operators would strengthen the generalization claim. Additionally, the gating mechanism currently relies on task-specific state-based rules; incorporating richer medical-domain constraints (e.g., anatomical boundaries and tissue-interaction models) remains an important direction for broader clinical applicability.

\bibliographystyle{IEEEtran}
\bibliography{references}

@article{iakovidis2025medical,
  title={Medical \& healthcare robotics: a roadmap for enhanced precision, safety, and efficacy},
  author={Iakovidis, Dimitris K and Vartholomeos, Panagiotis and Gall, Alexia Le and Cianchetti, Matteo and Ozioko, Oliver and Dahiya, Ravinder and Mazomenos, Evangelos and Vasconcelos, Francisco and Stoyanov, Danail and Philpott, Joe and others},
  journal={Measurement Science and Technology},
  volume={36},
  number={10},
  pages={103001},
  year={2025},
  publisher={IOP Publishing}
}

@article{wu2024review,
  title={A review on machine learning in flexible surgical and interventional robots: Where we are and where we are going},
  author={Wu, Di and Zhang, Renchi and Pore, Ameya and Dall’Alba, Diego and Ha, Xuan Thao and Li, Zhen and Zhang, Yao and Herrera, Fernando and Ourak, Mouloud and Kowalczyk, Wojtek and others},
  journal={Biomedical Signal Processing and Control},
  volume={93},
  pages={106179},
  year={2024},
  publisher={Elsevier}
}

@article{liu2024evolution,
  title={Evolution of surgical robot systems enhanced by artificial intelligence: a review},
  author={Liu, Yanzhen and Wu, Xinbao and Sang, Yudi and Zhao, Chunpeng and Wang, Yu and Shi, Bojing and Fan, Yubo},
  journal={Advanced Intelligent Systems},
  volume={6},
  number={5},
  pages={2300268},
  year={2024},
  publisher={Wiley Online Library}
}

@article{dagnino2024robot,
  title={Robot-assistive minimally invasive surgery: trends and future directions},
  author={Dagnino, Giulio and Kundrat, Dennis},
  journal={International Journal of Intelligent Robotics and Applications},
  volume={8},
  number={4},
  pages={812--826},
  year={2024},
  publisher={Springer}
}

@article{huang2023mri,
  title={MRI-guided robot intervention—current state-of-the-art and new challenges},
  author={Huang, Shaoping and Lou, Chuqian and Zhou, Ying and He, Zhao and Jin, Xuejun and Feng, Yuan and Gao, Anzhu and Yang, Guang-Zhong},
  journal={Med-X},
  volume={1},
  number={1},
  pages={4},
  year={2023},
  publisher={Springer}
}

@article{maier2017surgical,
  title={Surgical data science for next-generation interventions},
  author={Maier-Hein, Lena and Vedula, Swaroop S and Speidel, Stefanie and Navab, Nassir and Kikinis, Ron and Park, Adrian and Eisenmann, Matthias and Feussner, Hubertus and Forestier, Germain and Giannarou, Stamatia and others},
  journal={Nature Biomedical Engineering},
  volume={1},
  number={9},
  pages={691--696},
  year={2017},
  publisher={Nature Publishing Group UK London}
}

@article{moustris2011evolution,
  title={Evolution of autonomous and semi-autonomous robotic surgical systems: a review of the literature},
  author={Moustris, George P and Hiridis, Savvas C and Deliparaschos, Kyriakos M and Konstantinidis, Konstantinos M},
  journal={The international journal of medical robotics and computer assisted surgery},
  volume={7},
  number={4},
  pages={375--392},
  year={2011},
  publisher={Wiley Online Library}
}

@article{rivas2021review,
  title={A review on deep learning in minimally invasive surgery},
  author={Rivas-Blanco, Irene and Perez-Del-Pulgar, Carlos J and Garcia-Morales, Isabel and Munoz, Victor F},
  journal={IEEE Access},
  volume={9},
  pages={48658--48678},
  year={2021},
  publisher={IEEE}
}

@article{lee2024levels,
  title={Levels of autonomy in FDA-cleared surgical robots: a systematic review},
  author={Lee, Audrey and Baker, Turner S and Bederson, Joshua B and Rapoport, Benjamin I},
  journal={NPJ Digital Medicine},
  volume={7},
  number={1},
  pages={103},
  year={2024},
  publisher={Nature Publishing Group UK London}
}

@article{varghese2024artificial,
  title={Artificial intelligence in surgery},
  author={Varghese, Chris and Harrison, Ewen M and O’Grady, Greg and Topol, Eric J},
  journal={Nature medicine},
  volume={30},
  number={5},
  pages={1257--1268},
  year={2024},
  publisher={Nature Publishing Group US New York}
}

@article{vasey2023intraoperative,
  title={Intraoperative applications of artificial intelligence in robotic surgery: a scoping review of current development stages and levels of autonomy},
  author={Vasey, Baptiste and Lippert, Karoline AN and Khan, Danyal Z and Ibrahim, Mudathir and Koh, Chan Hee and Horsfall, Hugo Layard and Lee, Keng Siang and Williams, Simon and Marcus, Hani J and McCulloch, Peter},
  journal={Annals of Surgery},
  volume={278},
  number={6},
  pages={896--903},
  year={2023},
  publisher={LWW}
}

@article{kenig2024artificial,
  title={Artificial intelligence in surgery: a systematic review of use and validation},
  author={Kenig, Nitzan and Monton Echeverria, Javier and Muntaner Vives, Aina},
  journal={Journal of clinical medicine},
  volume={13},
  number={23},
  pages={7108},
  year={2024},
  publisher={MDPI}
}

@article{kim2024openvla,
  title={Openvla: An open-source vision-language-action model},
  author={Kim, Moo Jin and Pertsch, Karl and Karamcheti, Siddharth and Xiao, Ted and Balakrishna, Ashwin and Nair, Suraj and Rafailov, Rafael and Foster, Ethan and Lam, Grace and Sanketi, Pannag and others},
  journal={arXiv preprint arXiv:2406.09246},
  year={2024}
}

@article{black2024pi_0,
  title={{$\pi_0$}: A Vision-Language-Action Flow Model for General Robot Control},
  author={Black, Kevin and Brown, Noah and Driess, Danny and Esmail, Adnan and Equi, Michael and Finn, Chelsea and Fusai, Niccolo and Groom, Lachy and Hausman, Karol and Ichter, Brian and others},
  journal={arXiv preprint arXiv:2410.24164},
  year={2024}
}

@article{schmidgall2024general,
  title={General-purpose foundation models for increased autonomy in robot-assisted surgery},
  author={Schmidgall, Samuel and Kim, Ji Woong and Kuntz, Alan and Ghazi, Ahmed Ezzat and Krieger, Axel},
  journal={Nature Machine Intelligence},
  volume={6},
  number={11},
  pages={1275--1283},
  year={2024},
  publisher={Nature Publishing Group UK London}
}

@article{zargarzadeh2025decision,
  title={From decision to action in surgical autonomy: Multi-modal large language models for robot-assisted blood suction},
  author={Zargarzadeh, Sadra and Mirzaei, Maryam and Ou, Yafei and Tavakoli, Mahdi},
  journal={IEEE Robotics and Automation Letters},
  volume={10},
  number={3},
  pages={2598--2605},
  year={2025},
  publisher={IEEE}
}

@article{li2024deep,
  title={Deep learning for surgical workflow analysis: a survey of progresses, limitations, and trends},
  author={Li, Yunlong and Zhao, Zijian and Li, Renbo and Li, Feng},
  journal={Artificial Intelligence Review},
  volume={57},
  number={11},
  pages={291},
  year={2024},
  publisher={Springer}
}

@article{sone2023evolution,
  title={Evolution of a surgical system using deep learning in minimally invasive surgery},
  author={Sone, Kenbun and Tanimoto, Saki and Toyohara, Yusuke and Taguchi, Ayumi and Miyamoto, Yuichiro and Mori, Mayuyo and Iriyama, Takayuki and Wada-Hiraike, Osamu and Osuga, Yutaka},
  journal={Biomedical reports},
  volume={19},
  number={1},
  pages={45},
  year={2023},
  publisher={DA Spandidos}
}

@article{bilal2025temset,
  title={Temset-24k: Densely annotated dataset for indexing multipart endoscopic videos using surgical timeline segmentation},
  author={Bilal, Muhammad and Alam, Mahmood and Bapu, Deepashree and Korsgen, Stephan and Lal, Neeraj and Bach, Simon and Hajiyavand, Amir M and Ali, Muhammed and Soomro, Kamran and Qasim, Iqbal and others},
  journal={Scientific Data},
  volume={12},
  number={1},
  pages={1424},
  year={2025},
  publisher={Nature Publishing Group UK London}
}

@article{kolbinger2024artificial,
  title={Artificial Intelligence for context-aware surgical guidance in complex robot-assisted oncological procedures: An exploratory feasibility study},
  author={Kolbinger, Fiona R and Bodenstedt, Sebastian and Carstens, Matthias and Leger, Stefan and Krell, Stefanie and Rinner, Franziska M and Nielen, Thomas P and Kirchberg, Johanna and Fritzmann, Johannes and Weitz, J{\"u}rgen and others},
  journal={European Journal of Surgical Oncology},
  volume={50},
  number={12},
  pages={106996},
  year={2024},
  publisher={Elsevier}
}

@article{wagner2023comparative,
  title={Comparative validation of machine learning algorithms for surgical workflow and skill analysis with the HeiChole benchmark},
  author={Wagner, Martin and M{\"u}ller-Stich, Beat-Peter and Kisilenko, Anna and Tran, Duc and Heger, Patrick and M{\"u}ndermann, Lars and Lubotsky, David M and M{\"u}ller, Benjamin and Davitashvili, Tornike and Capek, Manuela and others},
  journal={Medical image analysis},
  volume={86},
  pages={102770},
  year={2023},
  publisher={Elsevier}
}

@article{maier2022surgical,
  title={Surgical data science--from concepts toward clinical translation},
  author={Maier-Hein, Lena and Eisenmann, Matthias and Sarikaya, Duygu and M{\"a}rz, Keno and Collins, Toby and Malpani, Anand and Fallert, Johannes and Feussner, Hubertus and Giannarou, Stamatia and Mascagni, Pietro and others},
  journal={Medical image analysis},
  volume={76},
  pages={102306},
  year={2022},
  publisher={Elsevier}
}

@article{li2025surgical,
  title={Surgical video workflow analysis via visual-language learning},
  author={Li, Pengpeng and Shu, Xiangbo and Feng, Chun-Mei and Feng, Yifei and Zuo, Wangmeng and Tang, Jinhui},
  journal={npj Health Systems},
  volume={2},
  number={1},
  pages={5},
  year={2025},
  publisher={Nature Publishing Group UK London}
}

\end{document}